\documentclass[11pt,a4paper]{article}
\usepackage[margin=0.83in]{geometry}
\usepackage[T1]{fontenc}
\usepackage[utf8]{inputenc}
\usepackage{lmodern}
\usepackage{microtype}
\usepackage{setspace}
\usepackage{amsmath,amssymb,booktabs,graphicx,array,tabularx}
\usepackage[font=small,labelfont=bf]{caption}
\usepackage[colorlinks=true,linkcolor=black,citecolor=blue!50!black,urlcolor=blue!50!black]{hyperref}
\usepackage{xcolor}
\begin{document}

\title{ContraRetrieve: Counterexample-Guided Evidence Retrieval for Faithful Chest Radiograph Reports}
\author{
Dan Smith$^{1}$, Rahul Patel$^{1}$\\[5pt]
\small $^{1}$Indian Institute of Technology Delhi,
New Delhi, India
}
\date{}
\maketitle
\begin{abstract}
\textbf{Background:} Retrieving similar radiology reports does not ensure that a generated report faithfully represents the patient image. We evaluated ContraRetrieve, a retrieval and verification framework that challenges proposed findings with counterexamples. \textbf{Methods:} The study used a 48,000-patient training/retrieval archive, separate validation and held-out partitions, and a 500-patient external evaluation set. We compared factual report errors, GREEN, and RadGraph F1 across five systems and examined error categories and design ablations. \textbf{Results:} ContraRetrieve reduced serious errors per report from 1.100 for the base generator to 0.700, while GREEN rose from 0.400 to 0.560 and RadGraph F1 from 0.277 to 0.352. In category-level adjudication, the total counted errors fell from 330 to 210, with 75 fewer false positive findings. Removing negative retrieval increased serious errors per report to 0.800. \textbf{Interpretation:} The findings associate counterexample-aware, spatially verified retrieval with fewer report-level factual errors in the evaluated setting, without establishing prospective clinical effectiveness.
\end{abstract}
\section{Introduction}
Generating a plausible report is not equivalent to reading an image correctly. A model may recover familiar radiology phrasing while inventing a pleural effusion, placing a lesion on the wrong side or omitting an important abnormality. Retrieval systems can supply similar reports, yet global similarity favors common templates and may reproduce precisely the kind of erroneous statements that ought to be checked. Clinical use requires a reason to believe each generated assertion is supported by the target study.

The key methodological question is how to use a retrieval archive to challenge, rather than merely support, proposed radiology statements.

We developed ContraRetrieve, an evidence-checking pipeline that uses positive and hard-negative retrieved examples as comparative references. The methodological hypothesis is that a counterexample should reduce confidence in a candidate assertion only when a target-image difference supports that conclusion. Retrieved reports are never accepted as ground truth for the current patient. The primary endpoint is the frequency of clinically significant factual errors per report, assessed independently of lexical similarity.
The difficulty is not confined to unusual diseases. A report may correctly identify an abnormality but assign it to the contralateral lung, or correctly describe a device while misrepresenting its position. These errors can coexist with otherwise accurate sentences, making a single report-level similarity score insufficient to locate the underlying problem. The relevant unit of verification is therefore the individual clinical proposition, considered in its anatomical and linguistic context.

Our study addresses three questions. First, does retrieval that explicitly includes contradictory cases improve factuality relative to conventional retrieval augmentation? Second, does conditioning verification on a localized region provide an advantage over global image--report similarity? Third, which categories of reporting error remain difficult after candidate assertions have been checked? The comparison and ablation results address these questions descriptively rather than establishing independent causal effects of the components.
\section{Related work}
Paired radiographs and reports are available through MIMIC-CXR and other cohorts, including CheXpert, ChestX-ray8, PadChest and VinDr-CXR \cite{mimic2019,chexpert2019,chestxray2017,padchest2020,vindr2022}. CheXlocalize adds region-level evidence against which retrieval-based explanations can be audited \cite{chexlocalize2022}. Contrastive image--text representations can organize report archives but may not detect subtle discrepancies in laterality or absence \cite{clip2021,convirt2022,gloria2021,medclip2022,chexzero2022,medklip2023,biomedclip2025}.

Report generators based on memory networks and generalist multimodal models provide strong candidate systems for an identical-input comparison \cite{r2gen2020,cmn2021,chexagent2024,radfm2023,roentgen2022}. Retrieval-augmented language generation offers a generic mechanism for supplying external context \cite{rag2020}; however, for radiology the provenance of that context must remain distinct from target-study evidence. Spatial prompting and region masks can be supplied by general or medical segmentation models \cite{sam2023,medsam2024}. Structured report entities and automated labelers support consistent error definitions \cite{radgraph2021,chexbert2020}.

Report-grounded segmentation offers localized evidence for evaluating individual textual findings \cite{xi2026grounding}, while image-focused retrieval reasoning provides a related approach to comparing visual referents \cite{xi2025taming}.

BLEU and embedding-based similarity metrics describe textual overlap or semantic relatedness \cite{bleu2002,bertscore2020}. Clinically informed measures such as RadCliQ and GREEN were proposed precisely because fluent reports can contain clinically consequential factual errors \cite{radcliq2023,green2024}. We therefore separate report readability from visual factuality throughout the study.

\subsection*{Retrieval as evidence rather than a substitute report}
An archive of prior cases offers examples of how findings are described, but it also introduces a potential source of error. A semantically similar report may describe a true finding for a different patient. Copying the description into the target report transfers language without transferring the evidence that justified it. This is a particularly important distinction for laterality, support devices and findings that are common in the retrieval collection. Retrieval therefore provides hypotheses and contrasts, not labels for the target examination.

Positive and counterexample retrieval have complementary roles. A positive example illustrates how an assertion may be expressed when its evidence is present. A counterexample identifies a similar context in which a particular feature is absent or differs in location. The retrieved images can only resolve the candidate assertion when the target study itself contains the relevant visual information. For this reason, the verification stage is separated from both retrieval ranking and report editing.
\section{Methods}
\subsection{Candidate assertions and counterexample retrieval}
ContraRetrieve operates on candidate assertions from a baseline radiology report generator. An assertion records the finding, polarity, site, and any severity or device qualifier. Separate retrieval streams identify reports and images that are compatible with the proposed assertion and examples that challenge it. Retrieval is constrained to the training archive and does not use evaluation-set reports as candidate exemplars.
\subsection{Assertion representation and retrieval constraints}
The candidate report is decomposed into propositions with explicit finding identity, polarity and anatomical qualifiers. This representation allows a verifier to compare like with like: an example of a left-sided effusion should not count as direct support for a right-sided claim, even if the two reports are otherwise similar. Retrieved positive and conflicting cases are maintained as separate sets so that global semantic similarity does not silently replace the role of visual contradiction.

The evaluation partitions are excluded from the retrieval archive. This matters because retrieving a target report, or a near-duplicate version of it, would make textual agreement look stronger without measuring the intended cross-case generalization. Within the training archive, retrieved examples remain conditional context for verification rather than adjudicated target-study truth.
\subsection{Clinical meaning of a counterexample}
A useful counterexample is not simply an unrelated examination. Its value arises from similarity in the surrounding clinical or anatomical context together with a difference in the proposition under consideration. For instance, an examination with a left pleural effusion can provide a contrast for a proposed right pleural effusion, whereas an unrelated normal examination provides little information about the localization problem. A negative example can also challenge the polarity of an assertion while leaving other findings unchanged. These are illustrative cases, not patient examples drawn from the evaluation cohort.

The distinction between contradiction and uncertainty is important. A report that does not mention a finding need not establish its absence; reporting practices, image quality, and the clinical indication can all influence what is documented. Similarly, a difference between two images does not by itself determine which of two assertions is correct. The interpretation of a retrieved negative case must consequently remain conditional on the evidence in the target radiograph.

\subsection{Visual support and constrained revision}
Let $q$ be a candidate assertion, $x$ the target radiograph, and $\mathcal R^{+},\mathcal R^{-}$ the supporting and conflicting retrieval sets. The support function compares local image evidence with the assertion and contrasting examples:
\begin{equation}
 s(q,x)=f_{\theta}\bigl(E_I(x),E_T(q),\mathcal R^+,\mathcal R^-,M(q,x)\bigr),
\end{equation}
where $M(q,x)$ is an assertion-conditioned visual region. A constrained editor retains assertions supported by target-image evidence and subjects unsupported or ambiguous assertions to revision or abstention. Retrieved language is treated as comparative evidence, not as an authoritative description of the target patient. The ablation variants remove negative examples, use global retrieval similarity alone, omit region verification, or use text-only editing.
\subsection{Decision scope and evidence attribution}
The framework separates three information sources: the draft report supplies candidate propositions, the archive supplies comparative cases, and the target radiograph supplies patient-specific evidence. This distinction limits the role of retrieval: even a close match cannot justify adding a finding solely because it is present in the retrieved report. Conversely, an abnormality visible in the target image should not be removed simply because it is absent from a retrieved case.

A conceptual representation of an assertion is $q=(f,p,a,u)$, where $f$ denotes the finding, $p$ its polarity, $a$ its anatomical location, and $u$ any additional qualifier. This notation describes the information used in verification rather than a claim about the exact software serialization. Under this representation, two propositions can agree on finding identity while disagreeing on polarity or location. The distinction is particularly relevant when comparing images that share multiple abnormalities.

A revision should preserve the meaning of unaffected assertions. In particular, removing an unsupported positive finding is different from affirmatively stating that the finding is absent. When a region cannot resolve the proposition, abstention or qualified language may be preferable to replacing one unsupported statement with another. The present evaluation does not separately quantify the accuracy or coverage of abstention.

\subsection{Population and evaluation}
The training/retrieval partition comprised 48,000 patients, 276,000 images, and 165,000 reports. Validation comprised 4,000 patients, 23,000 images, and 14,000 reports; internal evaluation comprised 5,000 patients, 30,000 images, and 18,000 reports. A separate external partition contained 500 patients, 500 images, and 500 reports. Images and reports are enumerated separately because one report may be associated with multiple radiographs.
\subsection{Outcomes and comparisons}
The main endpoint is clinically serious factual errors per report. GREEN and RadGraph F1 are reported as supplementary automated measures and do not replace clinical error assessment. The compared systems were the base generator, top-$k$ retrieval-augmented generation, assertion-aware retrieval, region-only verification, and ContraRetrieve. Separate error-class counts summarize false positive findings, omissions, laterality/location errors, and severity/device errors. We report observed aggregate results; confidence intervals and reviewer agreement cannot be inferred from category totals alone.
\subsection{Interpretation of the evaluation endpoints}
Serious factual errors per report is the primary outcome because the clinical consequences of a report depend on correctness of the statements it contains. GREEN and RadGraph F1 provide complementary evidence of report quality and entity--relation agreement. An improved lexical or graph-based metric cannot rule out a rare but consequential hallucination, so the error inventory is analyzed separately by category. The categorical counts and the report-normalized rates are retained on their respective scales; the former describe adjudicated instances, while the latter describe errors averaged over the evaluation unit.
\subsection{Scope of the quantitative analysis}
For a set of $N$ evaluated reports with serious-error counts $e_i$, the report-normalized error rate can be expressed as
\begin{equation}
\bar e=\frac{1}{N}\sum_{i=1}^{N}e_i.
\end{equation}
For a baseline rate $\bar e_b$ and a method rate $\bar e_m$, the descriptive relative reduction is $1-\bar e_m/\bar e_b$. This quantity is not a reduction in the probability of an erroneous report, since individual reports can contain more than one error. Likewise, the count of unsupported edits is an inventory of events rather than a normalized editing risk unless the total number of edits is known.

The cohort inventories do not establish the denominator used for each reported endpoint. In particular, separate internal and external performance estimates, the adjudicated report sample size, reviewer procedures, and the exact model and retrieval configurations are not specified in the available study record. The results below are therefore treated as aggregate descriptive comparisons; no significance test, confidence interval, or external generalization estimate is asserted.
\section{Results}
\subsection{Evaluation partitions}
The training/retrieval archive contained 48,000 patients, with 4,000 patients in validation, 5,000 in held-out internal evaluation, and 500 in external evaluation (Table~\ref{tab:p2cohort}).
\begin{table}[htbp]
\centering\small
\caption{Patient, radiograph, and report inventories by study partition.}\label{tab:p2cohort}
\setlength{\tabcolsep}{5pt}
\begin{tabular}{lrrr}\toprule
Partition & Patients & Images & Reports \\
\midrule
Training/retrieval archive & 48,000 & 276,000 & 165,000 \\
Validation & 4,000 & 23,000 & 14,000 \\
Held-out internal & 5,000 & 30,000 & 18,000 \\
External evaluation & 500 & 500 & 500 \\
\bottomrule\end{tabular}
\end{table}

\subsection{Clinical factuality and automated agreement}
\begin{figure}[htbp]\centering\includegraphics[width=\linewidth]{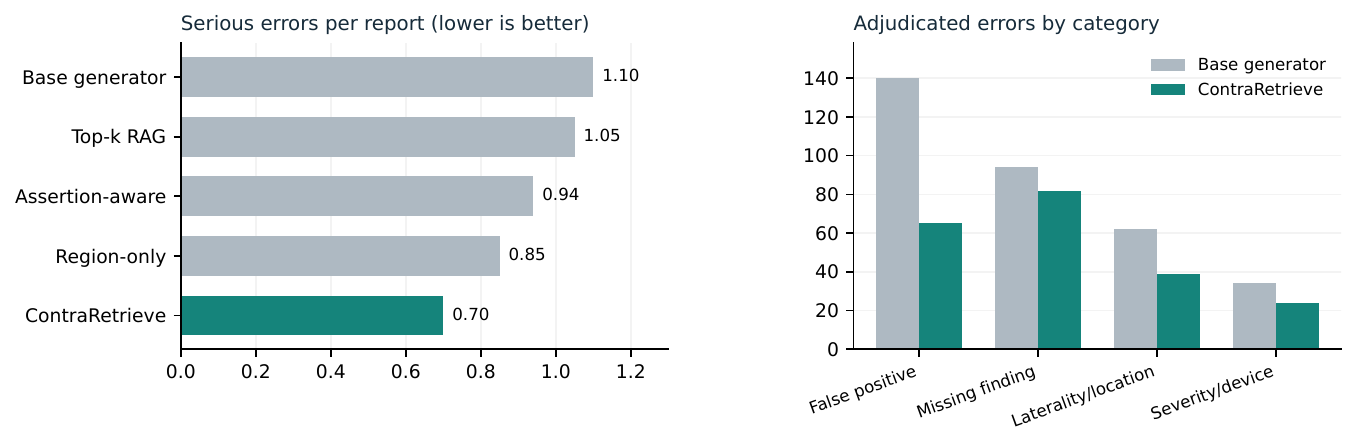}\caption{Report-level error rates and the separately tabulated clinical error categories from Tables~\ref{tab:p2main} and \ref{tab:p2errors}.}\label{fig:p2results}\end{figure}

The base generator produced 1.100 serious errors per report, compared with 0.700 for ContraRetrieve (Table~\ref{tab:p2main}), an absolute reduction of 0.400 or 36.4\% relative to the baseline. The best intermediate system, region-only verification, achieved 0.850; ContraRetrieve reduced the error rate by a further 0.150. GREEN improved from 0.400 to 0.560, and RadGraph F1 from 0.277 to 0.352, relative to the base generator. The systems are compared in Figure~\ref{fig:p2results}.
\begin{table}[htbp]
\centering\small
\caption{Report-level factuality and automated evaluation.}\label{tab:p2main}
\setlength{\tabcolsep}{5pt}
\begin{tabular}{lccc}\toprule
System & Serious errors/report $\downarrow$ & GREEN $\uparrow$ & RadGraph F1 $\uparrow$ \\
\midrule
Base generator & 1.100 & 0.400 & 0.277 \\
Top-$k$ RAG & 1.050 & 0.420 & 0.294 \\
Assertion-aware retrieval & 0.940 & 0.450 & 0.310 \\
Region-only verification & 0.850 & 0.490 & 0.324 \\
\textbf{ContraRetrieve} & \textbf{0.700} & \textbf{0.560} & \textbf{0.352} \\
\bottomrule\end{tabular}
\end{table}

\subsection{Distribution of factual errors}
The category-level review identified 330 errors with the base generator and 210 with ContraRetrieve, a reduction of 120 counted errors (Table~\ref{tab:p2errors}). False positive findings accounted for 75 of those 120 fewer errors (62.5\%). Omissions declined from 94 to 82; laterality/location errors from 62 to 39; and severity/device errors from 34 to 24. Category counts represent an adjudicated error inventory; they should not be converted to errors per report without its corresponding report denominator.
\begin{table}[htbp]
\centering\small
\caption{Adjudicated error counts and within-category differences for the base generator and ContraRetrieve.}\label{tab:p2errors}
\setlength{\tabcolsep}{5pt}
\begin{tabular}{lrrr}\toprule
Category & Baseline count & Full count & Count difference \\
\midrule
False positive finding & 140 & 65 & $-75$ \\
Missing finding & 94 & 82 & $-12$ \\
Laterality/location & 62 & 39 & $-23$ \\
Severity/device error & 34 & 24 & $-10$ \\
\textbf{Total} & \textbf{330} & \textbf{210} & $\mathbf{-120}$ \\
\bottomrule\end{tabular}
\end{table}

\subsection{Ablation of retrieval and verification}
Removing negative retrieval increased serious errors per report from 0.700 to 0.800 and doubled unsupported edits from 8 to 16. Global similarity alone resulted in 0.960 errors per report and 26 unsupported edits; removing region verification produced 0.920 and 28, respectively. Text-only editing had the highest error rate among the tested ablations (0.990) and the most unsupported edits (36; Table~\ref{tab:p2ablation}).
\begin{table}[htbp]
\centering\small
\caption{Ablation results. Unsupported edits are counts in the supplied evaluation inventory.}\label{tab:p2ablation}
\setlength{\tabcolsep}{5pt}
\begin{tabular}{lcc}\toprule
Variant & Serious errors/report $\downarrow$ & Unsupported edits $\downarrow$ \\
\midrule
\textbf{Full method} & \textbf{0.700} & \textbf{8} \\
No negative retrieval & 0.800 & 16 \\
Global similarity only & 0.960 & 26 \\
No region verification & 0.920 & 28 \\
Text-only editor & 0.990 & 36 \\
\bottomrule\end{tabular}
\end{table}
\subsection{Error reduction and failure modes}
The largest absolute reduction occurred for unsupported positive findings, which fell from 140 to 65. This category accounts for 75 of the 120 fewer counted errors. Laterality and location errors decreased by 23, while severity and device errors decreased by 10. The smaller reduction in missing findings, from 94 to 82, is important: verifying propositions present in a draft report more directly targets false statements than omissions. Detecting an omission may require a separate image-to-finding inventory rather than inspection of the sentences that a generator elected to write.

The ablations distinguish the contribution of hard negatives and spatial evidence. Removing negative retrieval raised serious errors per report from 0.700 to 0.800, whereas global similarity alone produced 0.960. Removing region verification yielded 0.920, and text-only editing yielded 0.990. These comparisons support the interpretation that contradictory retrieved cases are most useful when their differences can be resolved visually. The editor should not overwrite a supported target-image finding merely because a retrieved case lacks it.
\subsection{Magnitude and distribution of the observed changes}
Compared with conventional top-$k$ retrieval, ContraRetrieve reduced the serious-error rate from 1.050 to 0.700, a relative reduction of 33.3\%. Compared with region-only verification, the corresponding reduction was 17.6\%. GREEN increased by 0.160 and RadGraph F1 by 0.075 relative to the base generator. Although these measures move in the same favorable direction, they evaluate different aspects of report quality and should not be combined into a single clinical effect estimate.

The error inventory shows a 53.6\% reduction in false positive findings and a 12.8\% reduction in missing findings relative to their respective baseline counts. The difference between these reductions is consistent with the assertion-centered design: a proposition that appears in the draft can be challenged directly, whereas an omitted finding may never enter the verification stage. The remaining 210 counted errors also indicate that factual verification is incomplete. These reductions describe category totals and do not identify which individual cases improved or deteriorated.
\section{Discussion}
ContraRetrieve was associated with fewer report-level clinical errors than the base generator and the three intermediate retrieval or verification systems. The marked reduction in false positive findings suggests a particular benefit from challenging unsupported positive assertions, while the smaller reduction in omitted findings shows that verifying generated claims does not fully solve recognition of absent report content. The ablations provide complementary evidence that negative retrieval and region checks matter beyond text-only or globally similar references.

Radiographic world models investigate how evidence can be retained for both image understanding and generation \cite{xi2026radiographic}. Our results concern a different requirement: checking the factual support of a generated report against the specific source radiograph. The evaluation is retrospective and its results must not be equated with improved patient outcomes. GREEN and RadGraph F1 quantify aspects of reporting agreement but are not stand-alone measures of clinical safety. The category-level tabulation and the report-level rates are both retained without assuming their denominators are identical. Independent reader agreement, patient-level uncertainty estimates, error severity thresholds, and prospective workflow effects remain important for clinical interpretation.
\subsection{Why retrieval needs target-image verification}
Conventional retrieval can improve the availability of clinically appropriate terminology while also introducing statements that belong to another examination. ContraRetrieve addresses this tension by assigning a different role to retrieved evidence. A supporting example can help identify a plausible visual correlate, while a conflicting example can expose a feature that would otherwise be overlooked. Neither example resolves the current assertion without a comparison to the source image. The ablation findings are compatible with this interpretation, although the aggregate comparisons cannot establish whether the same mechanism explains every corrected report.

Spatial verification also has limitations. Diffuse abnormalities may not admit a compact region of interest, and the evidence for a negative finding may depend on inspecting a broader anatomical field. Overlapping structures, projection differences, and incomplete visualization can make an apparently contradictory case difficult to interpret. A region verifier should therefore be evaluated not only on its ability to localize visible findings, but also on its behavior when localization is ambiguous or the relevant anatomy is incompletely represented. These questions remain open in the present evaluation.

\subsection{Clinical interpretation and residual risk}
A lower mean error count does not establish that the method is safe for autonomous reporting. A system could reduce frequent low-impact errors while retaining uncommon errors with substantial clinical consequences. Nor does the reduction in false positives demonstrate that clinically important findings are never removed during revision. A future evaluation should distinguish corrected assertions, newly introduced errors, unchanged errors, and correctly preserved findings at the report level. Such an analysis would characterize the trade-off between correction and overcorrection more directly than aggregate counts.

The method is best interpreted as a candidate report-verification approach, rather than a replacement for professional image interpretation. The study does not evaluate turnaround time, reader workload, clinical decisions, or patient outcomes. These endpoints would require a separate clinical workflow study.

\subsection{Limitations and requirements for replication}
The available results do not identify the dataset source of every partition, the sampling criteria for adjudication, or the degree of overlap among dataset sources. Exclusion of evaluation reports from the retrieval index is necessary but not sufficient to demonstrate absence of leakage: patient identifiers, repeated examinations, and duplicate images should also be checked. Furthermore, the external cohort inventory alone does not establish transportability because separate external performance results are not available here.

Reproducibility depends on specifying the baseline generator and its checkpoint, the assertion extraction procedure, retrieval encoders and ranking rules, the numbers of positive and negative exemplars, region generation, verification thresholds, and the editor configuration. The clinical assessment also requires an operational definition of serious error, the number and qualifications of reviewers, and a documented procedure for disagreements. Without these details and patient-level results, independent replication and uncertainty estimation remain limited.

\section{Conclusion}
ContraRetrieve investigates counterexample-guided retrieval as a means of checking the factual content of generated chest radiograph reports. In the supplied aggregate evaluation, the method achieved fewer serious factual errors per report than the compared systems, with the largest category-level reduction occurring among false positive findings. Ablations suggest that both contradictory references and localized verification contribute to the observed differences. The smaller improvement in omissions and the remaining factual errors identify important limitations of assertion-centered verification. Further work requires a fully specified experimental protocol, independent clinical assessment, and separate evaluation of generalization before the approach can be considered for clinical deployment.

\section{Ethics, data availability and reproducibility}
This retrospective study used previously released, de-identified radiography resources for secondary computational analysis. No patients were newly recruited and no intervention or direct patient contact occurred. Source datasets were accessed under their applicable research-use terms, including credentialed access where applicable. The retrieval archive and evaluation partitions are reported separately to make potential leakage identifiable. Only aggregate comparisons are presented. Reproduction requires the original cohort partitions, retrieval-index membership, model configurations and clinical-error adjudication records.
\bibliographystyle{unsrt}
\bibliography{references}
\end{document}